\pdfoutput=1

\documentclass[11pt]{article}

\usepackage[final]{acl}

\usepackage{times}
\usepackage{latexsym}
\usepackage{amsmath}
\usepackage[T1]{fontenc}

\usepackage[utf8]{inputenc}

\usepackage{microtype}

\usepackage{inconsolata}

\usepackage{graphicx}
\usepackage{listings}
\usepackage{multirow}
\usepackage{float}
\usepackage{hyperref}
\title{A new approach for fine-tuning sentence transformers for intent classification and out-of-scope detection tasks}

\author{
  \textbf{Tianyi Zhang\textsuperscript{1,2,3}}\thanks{Work as a part of an internship at Cerence Inc.}, 
  \textbf{Atta Norouzian\textsuperscript{3}},
  \textbf{Aanchan Mohan\textsuperscript{1}},
  \textbf{Frederick Ducatelle\textsuperscript{3}} \\
  \textsuperscript{1}Northeastern University, Vancouver, BC, Canada, \\
  \textsuperscript{2}University of Victoria, Victoria, BC, Canada, \\
  \textsuperscript{3}Cerence Inc, Burlington, MA, USA,
\\
  \small{
    \href{mailto:tianyizhang2@uvic.ca}{tianyizhang2@uvic.ca
}, \href{mailto:atta.norouzian@cerence.com}{atta.norouzian@cerence.com}, \href{mailto:aa.mohan@northeastern.edu}{aa.mohan@northeastern.edu}, \href{frederick.ducatelle@cerence.com}{frederick.ducatelle@cerence.com}}
}

\begin{document}
\maketitle
\begin{abstract}
In virtual assistant (VA) systems it is important to reject or redirect user queries that fall outside the scope of the system. One of the most accurate approaches for out-of-scope (OOS) rejection is to combine it with the task of intent classification on in-scope queries, and to use methods based on the similarity of embeddings produced by transformer-based sentence encoders. Typically, such encoders are fine-tuned for the intent-classification task, using cross-entropy loss. Recent work has shown that while this produces suitable embeddings for the intent-classification task, it also tends to disperse in-scope embeddings over the full sentence embedding space. This causes the in-scope embeddings to potentially overlap with OOS embeddings, thereby making OOS rejection difficult. This is compounded when OOS data is unknown. To mitigate this issue our work proposes to regularize the cross-entropy loss with an in-scope embedding reconstruction loss learned using an auto-encoder. Our method achieves a 1-4\% improvement in the area under the precision-recall curve for rejecting out-of-sample (OOS) instances, without compromising intent classification performance.
\end{abstract}

\section{Introduction}
Virtual assistant (VA) systems often can handle only a limited scope of intents. Out-of-scope (OOS) rejection refers to the ability of a VA to identify and reject incoming queries that are outside its scope. This is a difficult \cite{fang2023} and increasingly important task in many scenarios. Our work is inspired by VAs in cars, which nowadays often operate in a hybrid mode where processing of certain user requests is handled locally, while others are transmitted to the cloud for response retrieval. Responding to users' requests using on-device/embedded models is cost-effective, quick, and, importantly, can safeguard sensitive information. Cloud models on the other hand are typically much bigger and can respond to a wider range of queries. In such a setting, it is important that the on-device natural language understanding (NLU) models not only identify user queries for intents that are in-scope but also accurately detect out-of-scope input so that they can be either routed to the cloud or ignored. Another important use case for OOS rejection is the combination of a light-weight, specialized VA that works tandem with large language models (LLMs) for free conversation with the user. Similar to the in-car use case, the specialized VA can be run \textit{before} the LLM and capture a subset of the incoming queries. This increases cost-effectiveness and controllability of the full solution, provided that it has good OOS rejection capabilities. 

The most common approach for intent classification while rejecting OOS samples is based on first generating an encoding for the sentences \cite{hendrycks2020,podolskiy2021revisiting} and then performing classification on them. In both \cite{hendrycks2020} and \cite{podolskiy2021revisiting} it was shown that the most suitable sentence encoders for this purpose are transformer-based encoders. Based on the task's domain, one could use one of the several sentence encoders available in the HuggingFace sentence transformer library \footnote{\url{https://sbert.net/}}. Fine-tuning sentence encoders on the domain-specific data leads to better intent classification accuracy. This fine-tuning typically is performed by applying a softmax to the sentence embeddings. At test time, the same softmax layer could be used to perform intent classification, however, the softmax tends to produce over-confident predictions even for OOS samples \cite{Dhamija2018, hendrycks2018}. Hence, after fine-tuning, the softmax layer is removed from the model and other classification approaches based on embedding similarities are used for intent classification and OOS rejection\cite{podolskiy2021revisiting}.

This fine-tuning approach is shown to be effective in teaching the model the class-discriminative features \cite{Fort2021} which in our task would result in a very good intent classification accuracy. However, fine-tuning without regularization could make the model forget some of the task-agnostic knowledge about general linguistic properties, which could help OOS detection \cite{Chen2023}. This shortcoming was tackled in \cite{zhou2021contrastive} by adding a regularization term based on contrastive loss. In this paper, we propose a new regularization term based on the global dispersion of in-scope sentence embeddings\footnote{Our code is available at : \url{https://github.com/SlangLab-NU/autoencoder-oos/tree/main}}. This is similar to the idea of deep one-class classification~\cite{ruff2018deep}, in which the model learns to project all in-scope samples into a relatively small neighborhood in the embedding space. In our approach, this is achieved by attaching an auxiliary autoencoder head to the fine-tuning architecture which reduces the global dispersion of the in-scpoe embeddings through minimization of reconstruction error. This approach is explained in detail in Section \ref{sec:method}.

\section{Related Work}
There are largely two categories of approaches for detecting OOS samples when performing intent-classification. The first category is based on explicitly teaching the model to distinguish between in-scope and OOS samples by introducing OOS samples during training. This is done by adding an extra OOS class to the classifier \cite{larson2019, qian-etal-2022-distinguish, choi2021outflip, zhan2021out} or by adding an auxiliary loss function to the cross entropy loss to enforce the model to output a uniform probability distribution over in-scope classes when dealing with OOS samples \cite{zheng2020}. These approaches only work if the OOS test samples are drawn from a distribution similar to that of the OOS training samples. In \cite{fang2023} the authors prove mathematically that it is not possible to detect samples outside of known distributions unless some conditions are met. This means for robust detection of OOS samples, the training OOS test samples have to represent a wide variety of possible distributions. While collecting such training samples is not feasible, synthesizing OOS samples using models like GANs \cite{ryu-etal-2018-domain,lee2018} and manifold learning \cite{goyal2020, Bhattacharya2023} have shown promise to make the decision boundary around in-system training samples as tight as possible.

The second category consists of approaches that rely only on in-scope training data without making any assumption about the OOS class. These approaches are largely based on sentence embeddings. Sentence embeddings generated by transformer encoders are shown to perform better than the ones generated using traditional NLU models \cite{hendrycks2020, podolskiy2021revisiting}. The classification of sentence embeddings into in-scope intent classes and into in-scope versus OOS could be done using non-parametric methods such as KNN \cite{zhou2022knn} or density based methods \cite{Chen2023, ren2021, xu-etal-2020-deep}. There is a trade-off between the model footprint and its accuracy when it comes to choosing between parametric and non-parametric approaches. Due to constraints on the size of the model put in the car we chose the parametric approach based on the Mahalanobis distance. 

The sentence embeddings could be generated using pretrained sentence transformers \cite{hendrycks2020} but fine-tuning the encoder for the task at hand provides more suitable embeddings \cite{darrin2024, zhou2021contrastive, Barnabo2023, zhou2022knn}. The work in \cite{zhou2021contrastive} highlights that while fine-tuning based on cross-entropy loss effectively separates sentence embeddings of different intent classes, it struggles to differentiate between in-scope samples and OOS samples. In that paper, this issue is tackled by adding a secondary loss function to the fine-tuning based on contrastive loss. The contrastive loss increases the distance between intent classes in the embedding space while reducing the distance between embeddings of the same intent class. However, since this loss tries to push the in-scope intent classes as far as possible from each other, the intent classes could start overlapping with OOS samples in the embedding space. Our approach inspired by the one-class classification in \cite{ruff2018deep} tries to reduce the dispersion of the in-scope intent classes in the embedding space by replacing the contrastive loss with reconstruction loss obtained using an autoencoder.    


\section{Methodology}
\label{sec:method}
This section discusses the details of our modelling formalism. Sub-section~\ref{sub:training} talks about our training cost-function(s), whereas sub-section~\ref{sub:inf} talks about our inference methodology.

\subsection{Model Fine-tuning}
\label{sub:training}

Figure~\ref{fig:model_arch} shows our model architecture. Let ${\mathbf{s}^i}$ denote the $d$ dimensional sentence embedding of the $i^{th}$ training sample generated after pooling the output of the transformer encoder. Here we use $\mathbf{y}^i$ to denote a $C$ dimensional one-hot vector associating $i^{th}$ input to one of $C$ in-scope intents. The $j^{th}$ element of $\mathbf{y}_i$ namely $y^{i}_{j}$ is equal to $1$ if and only if $\mathbf{s}^i$ belongs to the $j^{th}$ class where $ j \in \{1,\ldots,C\}$. In the baseline fine-tuning approach, a softmax layer is applied to ${\mathbf{s}^i}$ to map it to $\mathbf{e}^{i}$, a $C$-dimensional vector of probabilities. The cross-entropy loss $\mathcal{L}^{i}_{CE}$ of the $i^{th}$ training example is then calculated as:
\begin{equation}
\mathcal{L}^{i}_{CE} = -\sum_{j=1}^{C} y_{j}^{i} \log(e^{i}_{j})
\end{equation}
In the proposed fine-tuning approach, the sentence embedding ${\mathbf{s}^i}$ is passed to a second head which is comprised of an autoencoder network. The autoencoder reconstructs the embedding as $\mathbf{r}^{i}$. The reconstruction loss computed using mean-squared error is calculated as:
\begin{equation}
\mathcal{L}^{i}_{AE} = \frac{1}{d}\sum_{k=1}^{d}(s^{i}_{k}-r^{i}_{k})^{2}
\end{equation}
The architecture of the model along with the size of the layers of the autoencoder head are provided in Section~\ref{sub:pooling_and_dr}.
The final loss is calculated as follows weighted sum of the two losses described above as
\begin{equation}
\label{eq:cost_function}
    \mathcal{L}^{i} = (1-\alpha) \mathcal{L}^{i}_{CE} + \alpha \mathcal{L}^{i}_{AE}
\end{equation}
Here $\alpha$ tuned as a hyperparameter allows us to control the contribution of the individual losses towards the final loss. 

\subsection{Class-based Mean and Covariance Calculation}
\label{sub:cov}
After training, the autoencoder and the softmax heads are discarded. The transformer encoder trained with Eq.~\eqref{eq:cost_function} as the cost function is then primarily used for extracting sentence embeddings. Sentence embeddings using this transformer encoder are then generated for each training sample belonging to one of the $C$ in-scope intent classes. These per-class sentence embeddings are then used to construct a set of $C$ mean-vectors $\boldsymbol{\mu}_j$ where $j \in {1,\ldots,C}$. All of the training set sentence embeddings for the $C$ classes are then used to calculate a universal covariance matrix $\mathbf{\Sigma}$. 

\subsection{Classification and Inference}
\label{sub:inf}
For an incoming query $q$, if $\mathbf{s}^{q}$ is its corresponding sentence embedding, then the class-specific Mahalanobis distance $d_j$ is calculated as follows:
\begin{equation}
    d_j(\mathbf{s}^{q}) = \sqrt{(\mathbf{s}^{q} - \boldsymbol{\mu}_j)^\top \Sigma^{-1} (\mathbf{s}^{q} - \boldsymbol{\mu}_j)}
\end{equation}
Once the distances are calculated, a minimum distance $d_{min}(\mathbf{s}^{q})$ and the index $c_{min}(\mathbf{s}^{q})$ of the candidate centroid is picked as follows.
\begin{eqnarray}
    \label{eq:classification_1}
    d_{min}(\mathbf{s}^{q}) &=& \min_{j} d_j(\mathbf{s}^{q}) \\
    \label{eq:classification_2}
    c_{min}(\mathbf{s}^{q}) &=&  \arg\min_{j} d_j(\mathbf{s}^{q})
\end{eqnarray}
The quantity $d_{min}(\mathbf{s}^{q})$ is then compared to a threshold $\tau$ to determine if the query $q$ is in-scope or out-of-scope. This threshold is a hyper-parameter and is set empirically. If the query $q$ is determined to be in-scope then $c_{min}(\mathbf{s}^{q})$ is picked as the candidate class. This method of using a soft-max during training, but using the Mahalanobis distance during inference for classification is consistent with previous work~\cite{podolskiy2021revisiting, ren2021}.
\begin{figure*}[h]
    \centering
    \includegraphics[width=\linewidth]{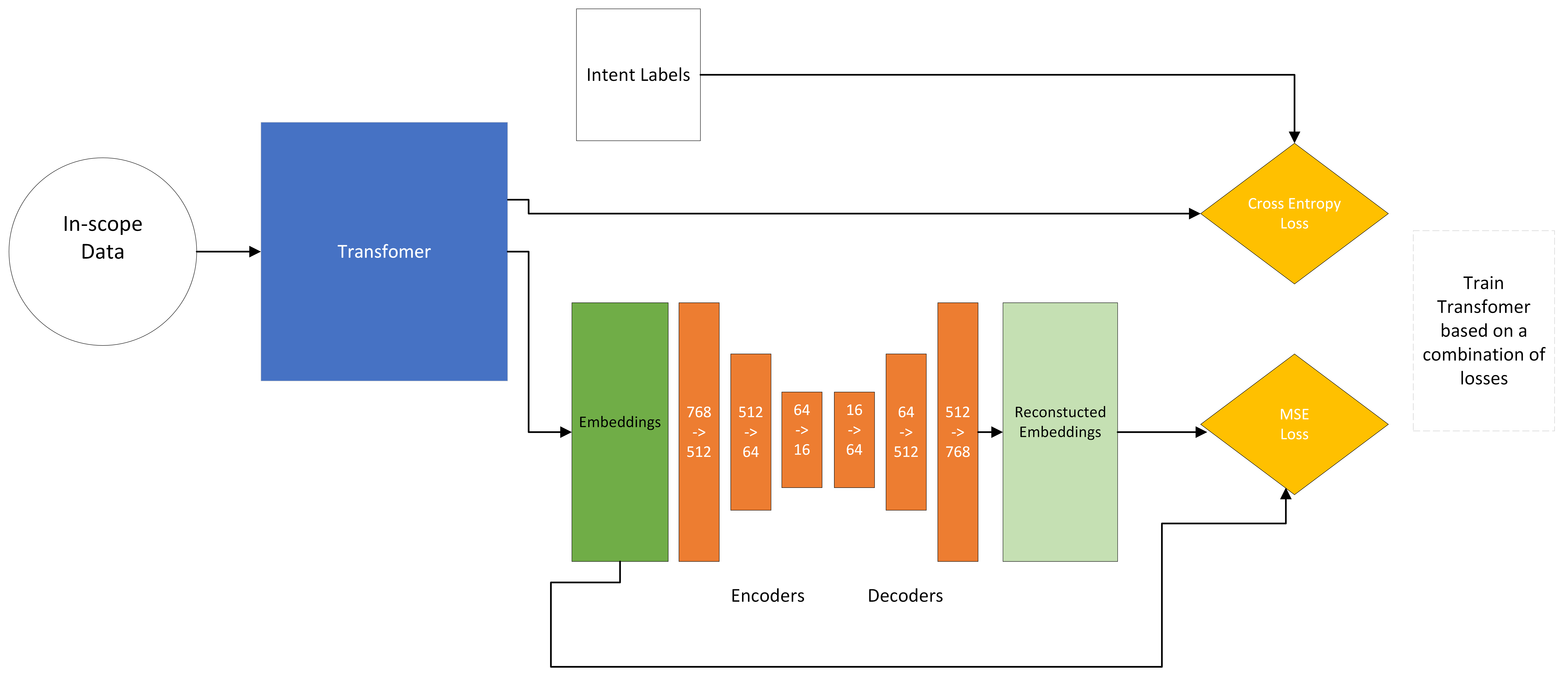}
    \caption{Model architecture for reducing the dispersion of in-scope embeddings. In-scope data is fine-tuned on a cross-entropy loss, with an auxiliary autoencoder loss.}
    \label{fig:model_arch}
\end{figure*}

\section{Experimental Setup}
This section talks about our experimental setup. The main objectives of our experimental setup is to evaluate the capability of our proposed fine-tuning approach to improve the model's ability to detect OOS queries robustly while maintaining in-scope intents classification accuracy.

\subsection{Sentence-encoder Configuration}
\label{sub:pooling_and_dr}
The \texttt{bert-base-uncased}~\cite{devlin2018bert} model followed by maxpooling was used to extract sentence embeddings.  Sentence embeddings from the transformer sentence encoder have dimensionality $d=768$. As shown in Figure~\ref{fig:model_arch} these embeddings pass through an autoencoder with a six-layer architecture designed to compress and reconstruct the sentence embeddings. The first 3 layers in the autoencoder reduce the data dimensionality from 768 to 512, 512 to 64, and finally from 64 to a 16-dimensional bottleneck.  The subsequent 3 layers reconstruct the sentence embedding back to its dimensionality of $d=768$. The transformer sentence-encoder is then trained using the objective function stated in Eq.~\eqref{eq:cost_function}. The errors are backpropagated from both heads back to the transformer sentence encoder. All layers of the transformer encoder model were fine-tuned.

\subsection{Hyperparameter Optimization and Training}
The training was found to be sensitive to the auto-encoder weight parameter $\alpha$. For this reason grid search was conducted for $\alpha$ with the following values $[0.01, 0.1, 0.2, 0.5, 0.9]$. The learning rate, batch size and no. of epochs were kept constant. It was found across the different validation sets that the optimal value for $\alpha=0.1$. The performance started to deteriorate drastically for higher values of $\alpha$.

The autoencoder weight $\alpha$ was then kept fixed for further hyperparameter optimization. Our work uses an open source hyperparameter optimization framework called Optuna \cite{optuna_2019}.  Learning rates between  \(1 \times 10^{-3}\) and \(5 \times 10^{-5}\) were explored using a logarithmic scale to prioritize smaller increments closer to the lower end of the spectrum, as transformer models often benefit from precise adjustments in learning rates. The number of training epochs ranged from 5 to 50. Batch size values were explored between 16, 32, 64, 128, 256, 512. The exact values for hyperparameters for each dataset appear in Appendix~\ref{app:hyp_param_values}.

\subsection{Evaluation Metrics}
\label{sub:eval_metrics}
The primary metric for assessing the effectiveness of our OOS detection was the Area Under the Precision-Recall curve (AUPR). It is important to mention that we label OOS samples as positive and in-scope samples as negative and hence we report AUPRood which signifies that. This metric is particularly suitable for comparing two binary classifiers when the test data is imbalanced like those with a high proportion of in-scope queries compared to OOS queries. The second metric is used is Area Under the ROC curve (AUROC). Our work additionally looks at the intent classification accuracy. This is important as our goal is to improve OOS rejection while maintaining in-scope intent classification accuracy.

\subsection{Datasets}

\textbf{CLINC150 Dataset:} The CLINC150 dataset~\cite{misc_clinc150_570} is a benchmark dataset for evaluating natural language understanding systems particularly in the context of intent and slot filling tasks. The data set comprises 150 intent classes with an extra class labeled as out-of-scope. The training data consists of 15,000 examples with 100 examples per intent. The out-of-scope intent was not used in training. The validation data consists of 3,000 examples with 20 examples per intent. The test data consists of 4,500 examples with 30 examples per intent. The data spans across 10 diverse domains, such as banking, credit cards, kitchen appliances making it comprehensive for real-world scenarios. Each data sample consists of a short text utterance, paired with an intent label.

\textbf{Stackoverflow Dataset:} This dataset is a curated subset from a challenge dataset originally published by Kaggle\footnote{\url{https://www.kaggle.com/c/predict-closed-questions-on-stack-overflow}}. The selection includes question titles that have been categorized into 20 distinct intent classes following the methodology proposed by Xu et al.~\cite{xu2017self}. Since this subset does not inherently include labeled out-of-scope (OOS) samples, we adopted the procedure described by Lin and Xu~\cite{lin-xu-2019-deep} to designate classes as either in-scope (IS) or OOS. Specifically, we retain classes that, combined, cover at least 75\% of the total dataset as IS. The remaining classes are considered OOS, and their instances are removed from the training dataset but retained and relabeled as OOS in the validation and test datasets. The specific details of dataset construction is detailed in Appendix~\ref{app:so_dataset}

\textbf{MTOP Dataset:} The MTOP dataset is a task-oriented dialogue dataset with a hierarchical structure of intent labels. In our experiments, we focus solely on the root-label of these intents. We utilized the English portion of this dataset, referred to as MTOP-EN, which comprises 87 intent classes in 11 domains. This dataset does not include a predefined out-of-scope (OOS) class. Based on the amount of data, the `timer' domain is chosen as the pre-defined OOS class. Our preprocessing filtered out in-scope (IS) domains with fewer than 10 occurrences per IS class. The in-scope data was then split into training, validation, and testing sets using a stratified approach based on intent labels to maintain an equal distribution of intents across these splits. We allocate OOS data between validation and testing sets, without stratification, due to the uniform label of OOS.

\textbf{Car Assistant Dataset:} This is an internal dataset. Due to it's original massive size, we randomly selected around 200,000 utterances used per run for training, validation and testing. This in-scope part of the dataset is derived from user interactions with car assistant systems and contains 46 distintc intent classes while. The OOS part is constructed from 14 different setes including sms messages, dictated emails, book snippets, tweets, internet-scraped text and some other unsupported text phrases.

\begin{table*}[h!]
\centering
\resizebox{\textwidth}{!}{
\begin{tabular}{|c|p{1.1cm}|p{2.2cm}|c|c|c|c|}
\hline
 &  &  &  &  &  & \textbf{Intent} \\
\textbf{Dataset} & \textbf{\#Train} & \textbf{\#Test(is/oos)} & \textbf{Fine-tuning} & \textbf{AUPRoos}& \textbf{AUROC} & \textbf{Classification} \\
 & & & & & & \textbf{Accuracy} (\%) \\
\hline
\multirow{2}{*}{CLINC150} & \multirow{2}{*}{15,000} & \multirow{2}{*}{4,500 / 1,000} & CE & 0.916 $\pm$ 0.007 & 0.977 $\pm$ 0.001 & 95.8 \\
 &  &  & CE+AE & 0.918 $\pm$ 0.004 & 0.978 $\pm$ 0.004 & 95.8 \\
\hline
\multirow{2}{*}{StackOverflow} & \multirow{2}{*}{79,048} & \multirow{2}{*}{16,940 / 14,617} & CE & 0.822 $\pm$ 0.053 & 0.881 $\pm$ 0.028 & 91.2 \\
 &  &  & CE+AE & \textbf{0.849 $\pm $ 0.050} & \textbf{0.893 $\pm$ 0.030} & 90.9 \\
\hline
\multirow{2}{*}{MTOP} & \multirow{2}{*}{14,465} & \multirow{2}{*}{4,134 / 997} & CE & 0.869 $\pm$ 0.018 & 0.974 $\pm$ 0.004 & 97.0 \\
 &  &  & CE+AE & \textbf{0.899 $\pm$ 0.039} & \textbf{0.979 $\pm$ 0.009} & 97.0 \\
\hline
\multirow{2}{*}{Car Assistant} & \multirow{2}{*}{600k} & \multirow{2}{*}{150k / 200k} & CE & 0.954 $\pm$ 0.005 & 0.959 $\pm$ 0.002 & 96.5 \\
 &  &  & CE+AE & \textbf{0.965 $\pm$ 0.004} & \textbf{0.966 $\pm$ 0.003} & 96.6 \\
\hline
\end{tabular}}
\caption{Comparison of cross-entropy (CE) fine-tuning and versus the joint cross-entropy and autoencoder objective (CE+AE). Here AUPRoos refers to the AUPR metric treating the OOS class as the positive class in the test set. The last column shows the intent classification accuracy result as a percentage.}
\label{table:performance_metrics}
\end{table*}

\section{Results and Discussion}
The OOS detection performance and intent classification accuracy of both the baseline and the proposed fine-tuning approaches are presented in Table \ref{table:performance_metrics}. The table has 4 rows and 7 columns. Each row of Table \ref{table:performance_metrics} contains results on one particular dataset. The first three columns show dataset name and a summary of numbers of utterances in each dataset. The fourth column shows the fine-tuning cost function used namely cross-entropy (CE), versus the joint cross-entropy and autoencoder (CE+AE) fine-tuning objective introduced in Eq.~\eqref{eq:cost_function}. The next three columns display our results for the evaluation metrics mentioned in Section~\ref{sub:eval_metrics}. As mentioned in the table caption, AUPRoos refers to calculating the AUPR by treating the OOS class in the test set as the positive class. AUROC refers the area under the receiver operating curve, and accuracy refers to intent classification accuracy using Eqs.~\eqref{eq:classification_1} and ~\eqref{eq:classification_2}. The intent classification accuracy is expressed as a percentage. 

The results show that in 3 out of 4 datasets we tested, the proposed method improved OOS detection, while maintaining the same in-scope intent classification accuracy. Specifically, the relative improvement with regard to AUPRoos is seen to be 3.22\% on the StackOverflow dataset, 3.45\% on the MTOP dataset and 1.15\% on the Car Assistant dataset. Due to its larger test set, the improvement on our internal car assistant dataset is statistically more significant than the improvement on the other two test sets. This can be attributed to the presence of a larger training set, which enables the autoencoder head to exert greater influence over the more than 100 million parameters of the sentence encoder.

\subsection{Embedding Dispersion}
\begin{table}[h!]
    \centering
\begin{tabular}{|c|c|c|}
\hline
\textbf{Dataset} & \textbf{CE} & \textbf{CE+AE} \\
\hline
CLINC150 & 17.767 & 17.762 \\
\hline
StackOverflow & 16.854 & 16.026 \\
\hline
MTOP & 17.269 & 16.744 \\
\hline
\end{tabular}
    \caption{Dispersion of fine-tuned models}
    \label{tab:dispersion}
\end{table}

We also measured the dispersion of the sentence embedding vector after baseline fine-tuning and after our proposed fine-tuning as shown in Table~\ref{tab:dispersion}. The dispersion was calculated as follows. For each training dataset that appears in Table~\ref{tab:dispersion}, training sentence embeddings were extracted first using our baseline cross-entropy (CE) model, and further using our model trained with joint cross-entropy and autoencoder objective (CE+AE). After extracting embeddings a global covariance matrix was calculated in each case namely $\boldsymbol{\Sigma_{CE}}$ and $\boldsymbol{\Sigma_{CE+AE}}$. To measure dispersion, the trace of each of these matrices were calculated~\cite{johnson2002applied}. The dispersion values thus calculated appear in Table~\ref{tab:dispersion}. The dispersion values illustrate that global dispersion of in-scope embeddings is smaller when our proposed fine-tuning is applied. It can be observed that the smaller the dispersion gets the higher OOS detection accuracy becomes when comparing the two fine-tuning approaches. This supports our argument that constraining the in-scope embeddings in a smaller neighborhood in the embedding space helps in the separation of in-scope and OOS samples. 

\subsection{Replacing the Model with a Large Language Model (LLM)}
Given the undeniable power of LLMs, one would naturally wonder what if the classification pipeline based on sentence encoder was replaced by an LLM. In other words, how well would a LLM  perform intent classification and OOS detection tasks without fine-tuning and just by prompt engineering. To answer this question we examined the performance of ChatGPT's gpt-3.5-turbo-0125 model from OpenAI on the MTOP dataset. We evaluated the performance of the LLM for intent-classification and OOS detection separately with different prompts as we noticed that if we ask the LLM to do both tasks, it will overwhelmingly classify most samples as OOS. Furthermore, due to the limitations in context size, we were limited to use 200 training examples but we made sure that there is at least one sample for each intent in the training set. In our setup, the system prompt is followed by the user prompt in which the model is provided with training sentences and a single test sentence. The exact system prompt is included in the Appendix~\ref{app:prompt}. Each experiment was repeated five times and the mean values of AUPR and AUROC as well as classification precision are presented in Table~\ref{tab:benchmark}.

\begin{table}[h!]
    \centering
    \begin{tabular}{|l|c|}
        \hline
        \textbf{Metric} & \textbf{Value} \\
        \hline
        Average AUPR & 0.624 $\pm$ 0.00440 \\
        \hline
        AUROC & 0.642 \\
        \hline
        Intent Classification Acc.(\%) & 82.9 \\
        \hline
    \end{tabular}
    \caption{Benchmark results on GPT-3.5}
    \label{tab:benchmark}
\end{table}

It is worth noting that even with a very limited amount of training data the LLM does a good job of classifying 82.9\% of the samples correctly. However, detecting OOS samples just by looking at a few in-scope samples is proven to be a more difficult task even for the LLM. Although comparing the performance of our approach to the LLM performance for this task is not fair because the latter only saw a fraction of the training samples, it shows that one could not simply replace the classifier with an LLM and expect high intent classification and OOS detection accuracy.

\section*{Conclusion}
In this paper, we introduce a new approach to fine-tuning sentence transformers used for intent classification, to improve their ability to detect OOS samples. We showed that sentence embeddings generated from encoders fine-tuned using the proposed approach provide better separation between in-scope and OOS samples while maintaining the separation between intent classes.

\section*{Limitations}
A limitation of our approach is that it requires more than a few examples per intent class during fine-tuning to make a big enough impact on the sentence encoder to improve OOS detection. In other words, it is not suitable for few-shot learning. This can be seen in the results given in Table~\ref{table:performance_metrics} where the OOS accuracy stays the same for the CLINC150 dataset, where the ratio of samples to intent classes is much smaller than for the other datasets.
The proposed approach was not evaluated for compositional or compound queries that contain both in-scope and OOS elements. This was mainly because in most virtual assistant systems the multi-intent queries are first broken into single-intent phrases, and then the classification step is performed. In addition, there are not many studies in the literature on this use case and not having publicly available datasets with such queries in them would make it difficult for us to benchmark our approach against SOTA approaches.

\section*{Acknowledgments}
The authors would like to acknowledge the financial support received from the Khoury West Coast Research Fund from the Khoury College of Computer Sciences at Northeastern University to support Tianyi as a Master's student. The work was continued as a PhD student internship co-funded by Cerence Inc and the MITACS Accelerate Program through the University of Victoria. The authors would like to acknowledge support from Dr. Yvonne Coady from the University of Victoria and Christian Marschke at Cerence Inc.
\bibliography{custom}

\begin{thebibliography}{29}
\providecommand{\natexlab}[1]{#1}

\bibitem[{mis(2020)}]{misc_clinc150_570}
 2020.
\newblock {CLINC150}.
\newblock UCI Machine Learning Repository.
\newblock {DOI}: https://doi.org/10.24432/C5MP58.

\bibitem[{Akiba et~al.(2019)Akiba, Sano, Yanase, Ohta, and Koyama}]{optuna_2019}
Takuya Akiba, Shotaro Sano, Toshihiko Yanase, Takeru Ohta, and Masanori Koyama. 2019.
\newblock Optuna: A next-generation hyperparameter optimization framework.
\newblock In \emph{Proceedings of the 25th {ACM} {SIGKDD} International Conference on Knowledge Discovery and Data Mining}.

\bibitem[{Barnabo et~al.(2023)Barnabo, Uva, Pollastrini, Rubagotti, and Bernardi}]{Barnabo2023}
Giorgio Barnabo, Antonio Uva, Sandro Pollastrini, Chiara Rubagotti, and Davide Bernardi. 2023.
\newblock \href {https://www.amazon.science/publications/supervised-clustering-loss-for-clustering-friendly-sentence-embeddings-an-application-to-intent-clustering} {Supervised clustering loss for clustering-friendly sentence embeddings: An application to intent clustering}.
\newblock In \emph{IJCNLP-AACL 2023}.

\bibitem[{Bhattacharya et~al.(2023)Bhattacharya, Gandhi, Huddar, S, Moroney, Saroop, and Bhagat}]{Bhattacharya2023}
Arindam Bhattacharya, Ankit Gandhi, Vijay Huddar, Ankith~M S, Aayush Moroney, Atul Saroop, and Rahul Bhagat. 2023.
\newblock \href {https://www.amazon.science/publications/beyond-hard-negatives-in-product-search-semantic-matching-using-one-class-classification-smocc} {Beyond hard negatives in product search: Semantic matching using one-class classification (smocc)}.
\newblock In \emph{WSDM 2023}.

\bibitem[{Chen et~al.(2023)Chen, Yang, Bi, and Sun}]{Chen2023}
Sishuo Chen, Wenkai Yang, Xiaohan Bi, and Xu~Sun. 2023.
\newblock Fine-tuning deteriorates general textual out-of-distribution detection by distorting task-agnostic features.
\newblock In \emph{EACL 2023}, page 564–579.

\bibitem[{Choi et~al.(2021)Choi, Shin, Kim, and Shin}]{choi2021outflip}
DongHyun Choi, Myeong~Cheol Shin, EungGyun Kim, and Dong~Ryeol Shin. 2021.
\newblock Outflip: Generating out-of-domain samples for unknown intent detection with natural language attack.
\newblock \emph{arXiv preprint arXiv:2105.05601}.

\bibitem[{Darrin et~al.(2024)Darrin, Staerman, Gomes, Cheung, Piantanida, and Colombo}]{darrin2024}
Maxime Darrin, Guillaume Staerman, Eduardo Dadalto~C{\^a}mara Gomes, Jackie~CK Cheung, Pablo Piantanida, and Pierre Colombo. 2024.
\newblock Unsupervised layer-wise score aggregation for textual ood detection.
\newblock In \emph{Proceedings of the AAAI Conference on Artificial Intelligence}, volume~38, pages 17880--17888.

\bibitem[{Devlin et~al.(2018)Devlin, Chang, Lee, and Toutanova}]{devlin2018bert}
Jacob Devlin, Ming-Wei Chang, Kenton Lee, and Kristina Toutanova. 2018.
\newblock Bert: Pre-training of deep bidirectional transformers for language understanding.
\newblock \emph{arXiv preprint arXiv:1810.04805}.

\bibitem[{Dhamija et~al.(2018)Dhamija, G{\"u}nther, and Boult}]{Dhamija2018}
Akshay~Raj Dhamija, Manuel G{\"u}nther, and Terrance~E. Boult. 2018.
\newblock \href {https://api.semanticscholar.org/CorpusID:53282534} {Reducing network agnostophobia}.
\newblock \emph{ArXiv}, abs/1811.04110.

\bibitem[{Fang et~al.(2023)Fang, Li, Lu, Dong, Han, and Liu}]{fang2023}
Zhen Fang, Yixuan Li, Jie Lu, Jiahua Dong, Bo~Han, and Feng Liu. 2023.
\newblock \href {https://arxiv.org/abs/2210.14707} {Is out-of-distribution detection learnable?}
\newblock \emph{Preprint}, arXiv:2210.14707.

\bibitem[{Fort et~al.(2021)Fort, Ren, and Lakshminarayanan}]{Fort2021}
Stanislav Fort, Jie Ren, and Balaji Lakshminarayanan. 2021.
\newblock \href {https://proceedings.neurips.cc/paper_files/paper/2021/file/3941c4358616274ac2436eacf67fae05-Paper.pdf} {Exploring the limits of out-of-distribution detection}.
\newblock In \emph{Advances in Neural Information Processing Systems}, volume~34, pages 7068--7081. Curran Associates, Inc.

\bibitem[{Goyal et~al.(2020)Goyal, Raghunathan, Jain, Simhadri, and Jain}]{goyal2020}
Sachin Goyal, Aditi Raghunathan, Moksh Jain, Harsha~Vardhan Simhadri, and Prateek Jain. 2020.
\newblock \href {https://arxiv.org/abs/2002.12718} {Drocc: Deep robust one-class classification}.
\newblock \emph{Preprint}, arXiv:2002.12718.

\bibitem[{Hendrycks and Gimpel(2018)}]{hendrycks2018}
Dan Hendrycks and Kevin Gimpel. 2018.
\newblock \href {https://arxiv.org/abs/1610.02136} {A baseline for detecting misclassified and out-of-distribution examples in neural networks}.
\newblock \emph{Preprint}, arXiv:1610.02136.

\bibitem[{Hendrycks et~al.(2020)Hendrycks, Liu, Wallace, Dziedzic, Krishnan, and Song}]{hendrycks2020}
Dan Hendrycks, Xiaoyuan Liu, Eric Wallace, Adam Dziedzic, Rishabh Krishnan, and Dawn Song. 2020.
\newblock \href {https://doi.org/10.18653/v1/2020.acl-main.244} {Pretrained transformers improve out-of-distribution robustness}.
\newblock In \emph{Proceedings of the 58th Annual Meeting of the Association for Computational Linguistics}, pages 2744--2751, Online. Association for Computational Linguistics.

\bibitem[{Johnson et~al.(2002)Johnson, Wichern et~al.}]{johnson2002applied}
Richard~Arnold Johnson, Dean~W Wichern, et~al. 2002.
\newblock Applied multivariate statistical analysis.

\bibitem[{Larson et~al.(2019)Larson, Mahendran, Peper, Clarke, Lee, Hill, Kummerfeld, Leach, Laurenzano, Tang, and Mars}]{larson2019}
Stefan Larson, Anish Mahendran, Joseph~J. Peper, Christopher Clarke, Andrew Lee, Parker Hill, Jonathan~K. Kummerfeld, Kevin Leach, Michael~A. Laurenzano, Lingjia Tang, and Jason Mars. 2019.
\newblock \href {https://doi.org/10.18653/v1/D19-1131} {An evaluation dataset for intent classification and out-of-scope prediction}.
\newblock In \emph{Proceedings of the 2019 Conference on Empirical Methods in Natural Language Processing and the 9th International Joint Conference on Natural Language Processing (EMNLP-IJCNLP)}, pages 1311--1316, Hong Kong, China. Association for Computational Linguistics.

\bibitem[{Lee et~al.(2018)Lee, Lee, Lee, and Shin}]{lee2018}
Kimin Lee, Honglak Lee, Kibok Lee, and Jinwoo Shin. 2018.
\newblock \href {https://arxiv.org/abs/1711.09325} {Training confidence-calibrated classifiers for detecting out-of-distribution samples}.
\newblock \emph{Preprint}, arXiv:1711.09325.

\bibitem[{Lin and Xu(2019)}]{lin-xu-2019-deep}
Ting-En Lin and Hua Xu. 2019.
\newblock \href {https://doi.org/10.18653/v1/P19-1548} {Deep unknown intent detection with margin loss}.
\newblock In \emph{Proceedings of the 57th Annual Meeting of the Association for Computational Linguistics}, pages 5491--5496, Florence, Italy. Association for Computational Linguistics.

\bibitem[{Podolskiy et~al.(2021)Podolskiy, Lipin, Bout, Artemova, and Piontkovskaya}]{podolskiy2021revisiting}
Alexander Podolskiy, Dmitry Lipin, Andrey Bout, Ekaterina Artemova, and Irina Piontkovskaya. 2021.
\newblock Revisiting mahalanobis distance for transformer-based out-of-domain detection.
\newblock In \emph{Proceedings of the AAAI Conference on Artificial Intelligence}, volume~35, pages 13675--13682.

\bibitem[{Qian et~al.(2022)Qian, Qi, Wang, Kunc, and Potdar}]{qian-etal-2022-distinguish}
Cheng Qian, Haode Qi, Gengyu Wang, Ladislav Kunc, and Saloni Potdar. 2022.
\newblock \href {https://doi.org/10.18653/v1/2022.emnlp-industry.51} {Distinguish sense from nonsense: Out-of-scope detection for virtual assistants}.
\newblock In \emph{Proceedings of the 2022 Conference on Empirical Methods in Natural Language Processing: Industry Track}, pages 502--511, Abu Dhabi, UAE. Association for Computational Linguistics.

\bibitem[{Ren et~al.(2021)Ren, Fort, Liu, Roy, Padhy, and Lakshminarayanan}]{ren2021}
Jie Ren, Stanislav Fort, Jeremiah Liu, Abhijit~Guha Roy, Shreyas Padhy, and Balaji Lakshminarayanan. 2021.
\newblock \href {https://arxiv.org/abs/2106.09022} {A simple fix to mahalanobis distance for improving near-ood detection}.
\newblock \emph{Preprint}, arXiv:2106.09022.

\bibitem[{Ruff et~al.(2018)Ruff, Vandermeulen, Goernitz, Deecke, Siddiqui, Binder, M{\"u}ller, and Kloft}]{ruff2018deep}
Lukas Ruff, Robert Vandermeulen, Nico Goernitz, Lucas Deecke, Shoaib~Ahmed Siddiqui, Alexander Binder, Emmanuel M{\"u}ller, and Marius Kloft. 2018.
\newblock Deep one-class classification.
\newblock In \emph{International conference on machine learning}, pages 4393--4402. PMLR.

\bibitem[{Ryu et~al.(2018)Ryu, Koo, Yu, and Lee}]{ryu-etal-2018-domain}
Seonghan Ryu, Sangjun Koo, Hwanjo Yu, and Gary~Geunbae Lee. 2018.
\newblock Out-of-domain detection based on generative adversarial network.
\newblock In \emph{Proceedings of the 2018 Conference on Empirical Methods in Natural Language Processing}, pages 714--718, Brussels, Belgium. Association for Computational Linguistics.

\bibitem[{Xu et~al.(2020)Xu, He, Yan, Liu, Liu, and Xu}]{xu-etal-2020-deep}
Hong Xu, Keqing He, Yuanmeng Yan, Sihong Liu, Zijun Liu, and Weiran Xu. 2020.
\newblock \href {https://doi.org/10.18653/v1/2020.coling-main.125} {A deep generative distance-based classifier for out-of-domain detection with mahalanobis space}.
\newblock In \emph{Proceedings of the 28th International Conference on Computational Linguistics}, pages 1452--1460. International Committee on Computational Linguistics.

\bibitem[{Xu et~al.(2017)Xu, Xu, Wang, Zheng, Tian, and Zhao}]{xu2017self}
Jiaming Xu, Bo~Xu, Peng Wang, Suncong Zheng, Guanhua Tian, and Jun Zhao. 2017.
\newblock Self-taught convolutional neural networks for short text clustering.
\newblock \emph{Neural Networks}, 88:22--31.

\bibitem[{Zhan et~al.(2021)Zhan, Liang, Liu, Fan, Wu, and Lam}]{zhan2021out}
Li-Ming Zhan, Haowen Liang, Bo~Liu, Lu~Fan, Xiao-Ming Wu, and Albert Lam. 2021.
\newblock Out-of-scope intent detection with self-supervision and discriminative training.
\newblock \emph{arXiv preprint arXiv:2106.08616}.

\bibitem[{Zheng et~al.(2020)Zheng, Chen, and Huang}]{zheng2020}
Yinhe Zheng, Guanyi Chen, and Minlie Huang. 2020.
\newblock Out-of-domain detection for natural language understanding in dialog systems.
\newblock \emph{IEEE/ACM Transactions on Audio, Speech, and Language Processing}, 28:1198--1209.

\bibitem[{Zhou et~al.(2021)Zhou, Liu, and Chen}]{zhou2021contrastive}
Wenxuan Zhou, Fangyu Liu, and Muhao Chen. 2021.
\newblock Contrastive out-of-distribution detection for pretrained transformers.
\newblock In \emph{Proceedings of the 2021 Conference on Empirical Methods in Natural Language Processing}, pages 1100--1111.

\bibitem[{Zhou et~al.(2022)Zhou, Liu, and Qiu}]{zhou2022knn}
Yunhua Zhou, Peiju Liu, and Xipeng Qiu. 2022.
\newblock Knn-contrastive learning for out-of-domain intent classification.
\newblock In \emph{Proceedings of the 60th Annual Meeting of the Association for Computational Linguistics (Volume 1: Long Papers)}, pages 5129--5141.

\end{thebibliography}

\appendix

\section{Appendix}
\label{sec:appendix}

\subsection{Details of dataset construction}
\label{app:dataset}
\subsubsection{Stackoverflow dataset}
\label{app:so_dataset}
The dataset is divided into training, validation, and testing sets using a stratified split approach. The stratification ensures that the relative frequency of IS and OOS labels is maintained across the splits. This procedure is replicated across five different IS-OOS class configurations (splits), each initiated with a unique random seed for repeatability. For each split, the dataset undergoes:
\begin{enumerate}
    \item Filtering to include only the specific 20 categories from the original dataset. The labels selected for inclusion in this subset are as follows: `svn', `oracle', `bash', `apache', `excel', `matlab', `cocoa', `visual-studio', `osx', `wordpress', `spring', `hibernate', `scala', `sharepoint', `ajax', `drupal', `qt', `haskell', `linq', `magento'.
    \item Random shuffling and selection of tags to meet the 75\% threshold for IS designation.
    \item Out-of-domain data, not meeting the IS criteria, is split equally into validation and test sets, labeled as OOS.
\end{enumerate}

\subsection{Exact hyper-parameter values obtained for various datasets}
\label{app:hyp_param_values}
\begin{table}[H]
\centering
\begin{tabular}{|l|c|c|c|c|}
\hline
\textbf{Dataset} & $\alpha$ & lr & bs & \# epochs \\
\hline
CLINC150 & 0.1 & $10^{-4}$ & 256 & 15\\
\hline
StackOverflow & 0.1 & 5 x $10^{-5}$ & 1024 & 6\\
\hline
MTOP & 0.1 & 2.25 x $10^{-5}$ & 128 & 10\\
\hline
Car Assistant & 0.1 & 2.25 x $10^{-5}$ & 1024 & 7\\
\hline
\end{tabular}
\caption{Parameter values for different datasets. Here $\alpha$ refers to the autoencoder importance, lr refers to the learning rate, bs refers to the batch size and \# epochs refers to the number of epochs.}
\label{table:params}
\end{table}

\subsection{Prompt given to ChatGPT 3.5}
\label{app:prompt}
In order to evaluate our approach against ChatGPT 3.5 we used the following system prompt for OOS detection task: 
\begin{center}
\begin{lstlisting}
You are an AI assistant specialized in natural language processing tasks. You will be provided with training samples consisting of sentences and their corresponding intents. Your task is to determine whether a given sentence is in-scope (belongs to a known intent) or out-of-scope (does not belong to any known intent). Based on the provided training data, classify each input sentence and return a JSON object indicating whether the sentence is in-scope or out-of-scope. If the sentence is in-scope, also provide the intent name. If the sentence is out-of-scope, indicate that it is out-of-scope. The in-scope intents must match exactly with the intents provided in the training data except for oos. Instructions: 1. For each input sentence, determine if it is in-scope or out-of-scope based on the provided training data. 2. If the sentence is in-scope, return a JSON object with { inscope: true, scope: "intent_name" }. The intent name must match exactly with the intents provided in the training data. 3. If the sentence is out-of-scope, return a JSON object with { inscope: false, scope: "oos" }. 
\end{lstlisting}
\end{center}

System Prompt for classification task:
\begin{lstlisting}
You are an AI assistant specialized in natural language processing tasks. You will be provided with training samples consisting of sentences and their corresponding intents. Your task is to classify a given sentence's intent. Based on the provided training data, classify the input sentence and return a JSON object indicating the intent of the sentence. The intents must match exactly with the intents provided in the training data. Return a JSON object with { intent: "intent_name" }.
\end{lstlisting}
\end{document}